\documentclass[letterpaper, 10 pt, conference]{ieeeconf}  

\IEEEoverridecommandlockouts                              
\usepackage{color}
\usepackage{graphics} 
\usepackage{amsmath} 
\usepackage{graphicx}
\usepackage{algorithmicx}
\usepackage{algpseudocode,algorithm,algorithmicx}
\title{\LARGE \bf
Multi-stage Suture Detection for Robot Assisted Anastomosis based on Deep Learning 
}

\author{Yang Hu$^{1}$*, Yun Gu$^{2}$*, Jie Yang$^{2}$ and Guang-Zhong Yang$^{1}$
\thanks{*    Equal Contribution}
\thanks{$^{1}$Yang Hu and Guang-Zhong Yang are with the Hamlyn Centre for Robotic Surgery, Imperial College London, SW72AZ, UK}%
\thanks{$^{2}$Yun Gu and Jie Yang are with School of Biomedical Engineering and Institute of Image Processing \& Pattern Recognition, Shanghai Jiao Tong University, Shanghai, CHINA. Yun Gu is also with the Hamlyn Centre for Robotic Surgery, Imperial College London, SW72AZ, UK}%
}

\begin{document}
\maketitle
\thispagestyle{empty}
\pagestyle{empty}
\begin{abstract}
In robotic surgery, task automation and learning from demonstration combined with human supervision is an emerging trend for many new surgical robot platforms. One such task is automated anastomosis, which requires bimanual needle handling and suture detection. Due to the complexity of the surgical environment and varying patient anatomies, reliable suture detection is difficult, which is further complicated by occlusion and thread topologies. In this paper, we propose a multi-stage framework for suture thread detection based on deep learning. Fully convolutional neural networks are used to obtain the initial detection and the overlapping status of suture thread, which are later fused with the original image to learn a gradient road map of the thread. Based on the gradient road map, multiple segments of the thread are extracted and linked to form the whole thread using a curvilinear structure detector. Experiments on two different types of sutures demonstrate the accuracy of the proposed framework.  
\end{abstract}

\section{INTRODUCTION}

Detection of curvilinear structure is a ubiquitous but challenging task. Typical examples include road extraction from satellite images \cite{jin2005automated}, tracking guide wire/catheter in fluorescence images \cite{wang2009robust} and locating blood vessels in retinal images \cite{hoover2000locating}. In this paper, we focus on the detection of a special curvilinear object: the suture thread. Suturing is a fundamental step in most surgical procedures and for automated anastomosis with surgical robots, effective thread manipulation is essential. The technique of robust suture  detection is vital in many applications including trainee suturing skill evaluation, suture augmentation for robotic-assisted surgery and thread recognition for automatic suturing.

The task for thread detection is to reconstruct 2D or 3D models based on observed images which reflects the geometry structure of the threads. This reconstruction is however challenging due to the following factors: (1) the deformation of threads can be large and self-occlusion makes it difficult to recover all segments of threads; (2) the operation environment is complicated in surgical settings that can easily lead to the occlusion with surgical tools and body tissues.
. 
\begin{figure}[!t]
\centering
\includegraphics[width=0.47\textwidth]{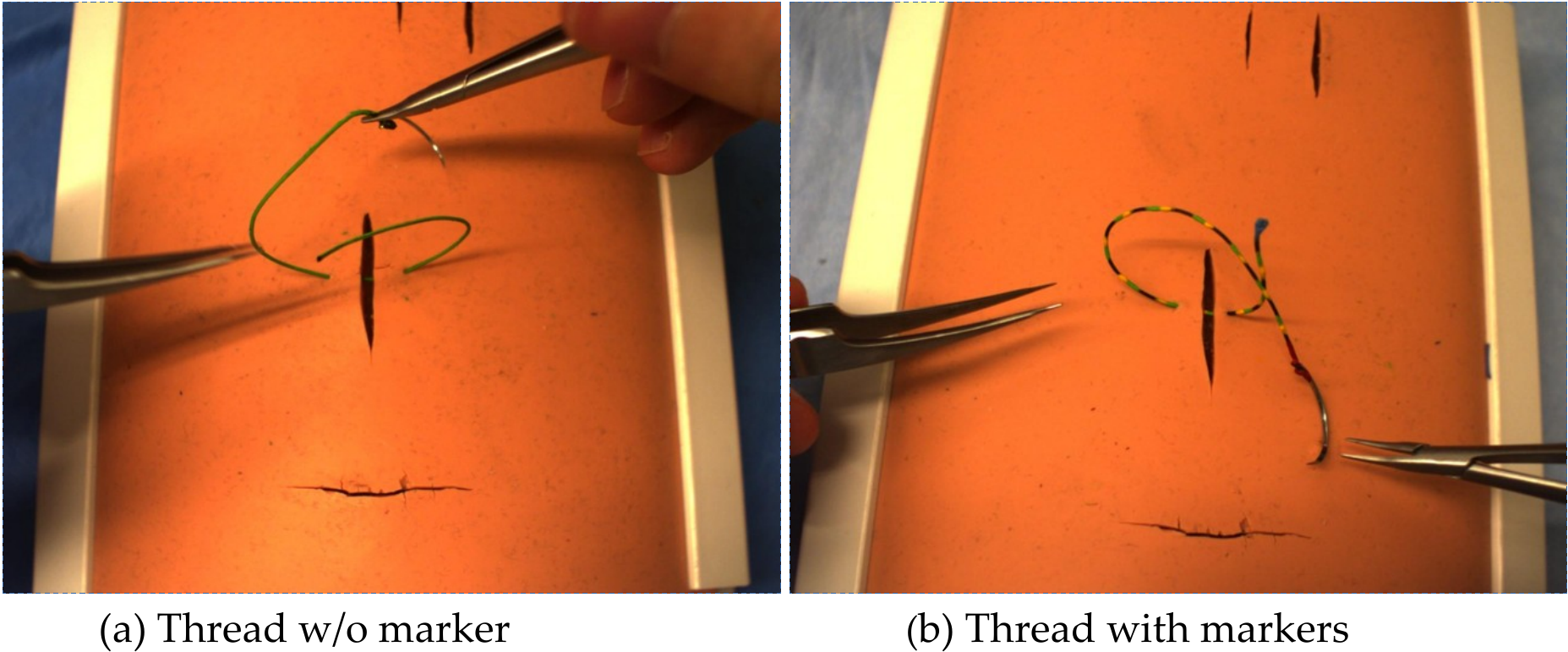}
\caption{Two types of threads used in the experiments}\label{fig::marker}
\end{figure}
\begin{figure*}[!t]
\centering
\includegraphics[width=0.95\textwidth]{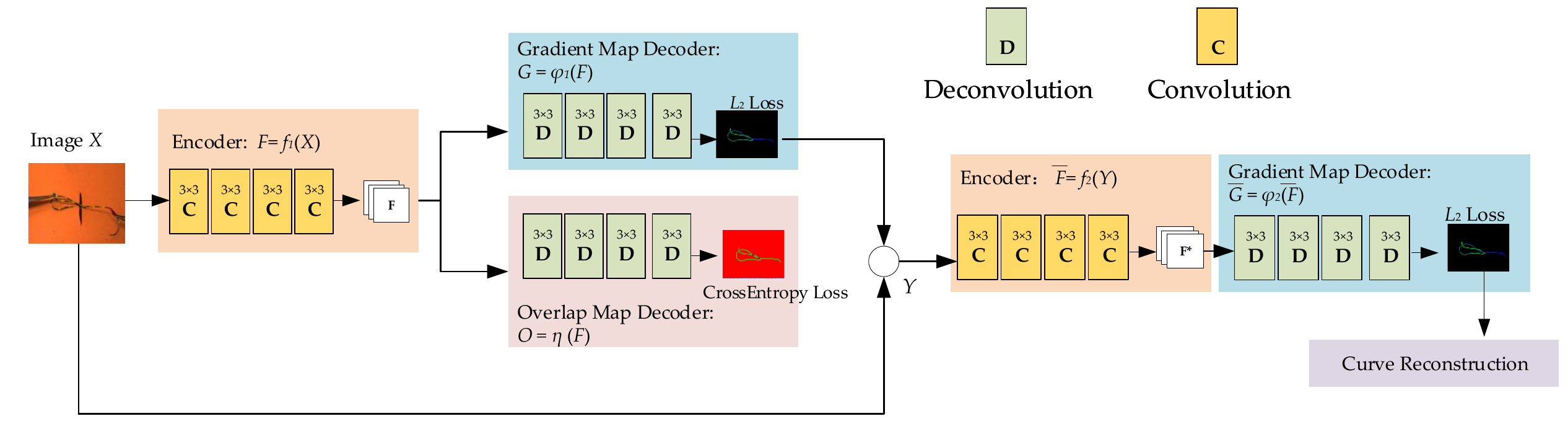}
\caption{The workflow of the proposed method.}\label{fig::workflow}
\end{figure*}

Recognition of a linear and deformable object, like a thread or rope, is a challenging task which attracts many researchers’ attention in the field of robotics. Many previous examples try to use RGB-D perception to obtain 3D points cloud of the thread, probabilistic model \cite{schulman2013tracking}, deformable registration \cite{schulman2016learning} and graph model \cite{lui2013tangled} have been applied to these recognition processes.  Due to the low resolution of RGB-D sensing, these methods using RGB-D perception only recognize a thick rope with diameter of several centimeters rather than a thin suturing wire less than 1 millimeter. Using model is a key aspect in tracking linear deformable object. These models can be complex physical model involving bending, twisting, tension energy, and length constraints, like \cite {javdani2011modeling}, or simple model like a spline. The tracking is accomplished by minimizing both the external energy (calculated by using the observed image and the model projection) and internal energy (e.g. bending, twisting energy). One limitation of using these models is that they may fail in the frequent situation that the thread suffers severe bending or entanglement. In addition, the tool-thread contact and thread-thread contact is too difficult to model. The use of temporal information is another interesting point in previous work, e.g. \cite {ambrosini2017fully} proposed that previous frames can help in detection of a twisted catheter in current frame. We human can untangle the complex structure of a curvilinear object in a single image without previous information; however, the temporal infomation may help in recognition the topological structure of the object and help in ambiguity case. Recently, there are works \cite {padoy2012deformable, jackson2015automatic, jackson2017real} focusing on surgical thread tracking, both using RGB stereo camera system. These two methods are based on similar idea even though their detailed methods are different. A 3D non-uniform rational B-spline is projected on to 2D image and an optimization on the spline parameters is performed to force the projection located on the observed thread. The spline is projected back to 3D once a good match is achieved. The method of optimization a spline model based on 2D observation can be called as a top-down approach. On the contrary, a bottom-up approach works by extracting the curvilinear features and later combine them to form a reasonable curve shape \cite{steger1998unbiased, strokina2013detection}. The top-down approach works well with simple tracking case without thread self-intersection or occlusion. However, as the author \cite{jackson2015automatic} concluded that their method does not handle well with segmentation outliers and unable to detect when the thread intersects itself (e.g. overhand knot).  On one hand, the poor performance on complex cases is because \cite{padoy2012deformable} did not implement any segmentation and \cite{jackson2015automatic, jackson2017real} use a segmentation method which is unable to exclude irrelevant curvilinear features. On the other hand, the projected 2D spline model may cover only a part of the observed thread rather than the whole thread, because there is no mechanism to indicate all the observed thread is covered by the 2D spline model.

Recently, the deep convolutional neural network (CNN) has boosted the segmentation tasks of curvilinear objects in medical applications including retina vessels ~\cite{fu2016retinal} and airways ~\cite{charbonnier2017improving}. In this paper, we propose a deep multi-stage detection (DMSD) framework for surgical thread detection, which uses a supervised, bottom-up and model-free approach. Instead of the isolated steps of segmentation and curve structure reconstruction in previous work, the proposed method tends to directly output the structural information (a gradient road map) of thread via deep convolutional neural networks (CNN). We firstly estimate the dense prediction of curvilinear structure and the overlapping status of threads by a two-branched CNN. The initial estimation is further fused with the original visual observation and fed into the second stage CNN to generate the refined prediction. Based on the gradient map, multiple segments of the thread are extracted and linked to form the whole thread with organized points located on its centerline using a curvilinear structure detector. Our method does not rely on any model for optimization, a simple spline model is used for smoothing the curve reconstruction at the last step.  Experiments are conducted on two types of surgical threads (with and without pattern) in a knot tying task as shown in Fig. 1. Experiment indicates that the proposed detection method can achieve sound performance when no occlusion or finite self-occlusion (e.g. overhand knot). For the tools-occlusion cases, the detection deteriorates for both type of threads, however, human level detection performance can be achieved for the thread with pattern.

\begin{figure}[t]
\centering
\includegraphics[width=0.45\textwidth]{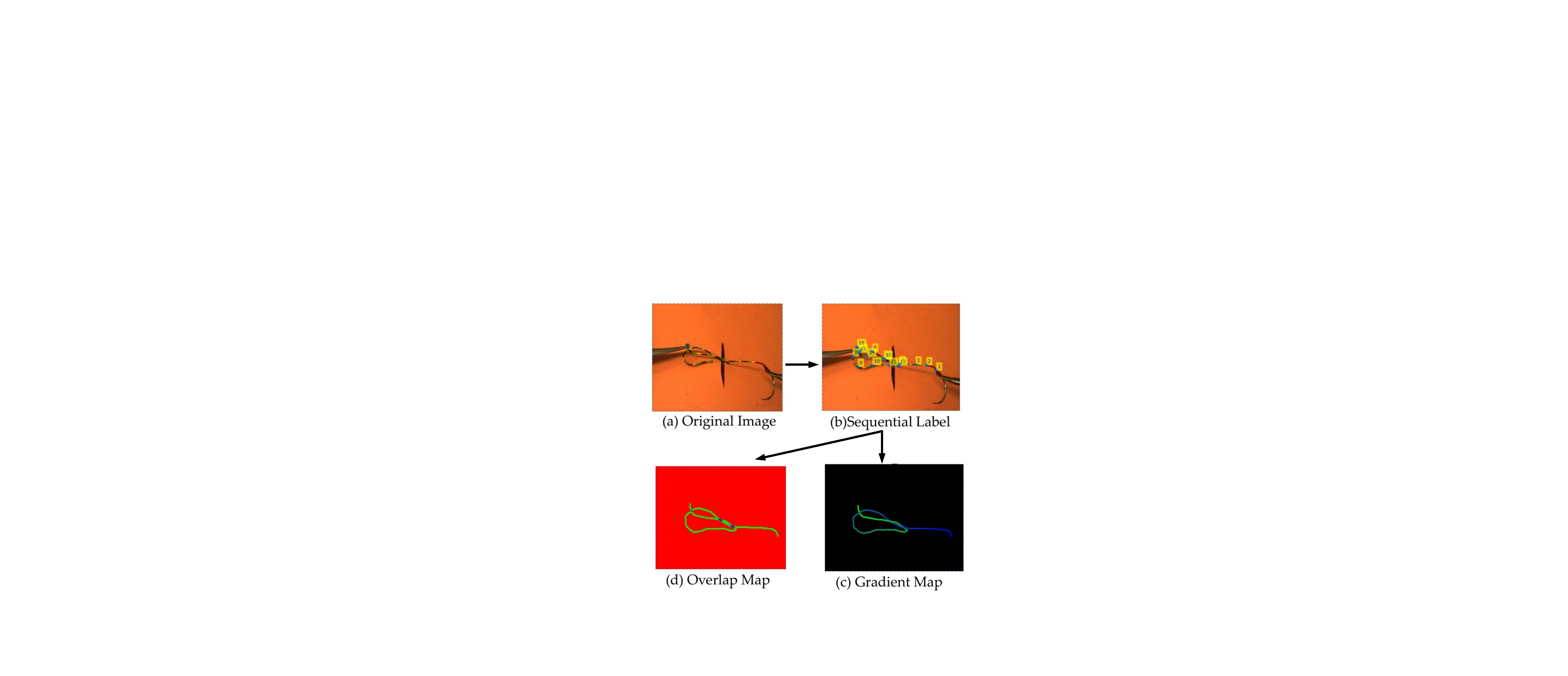}
\caption{Data collection and ground truth annotation. (a) A knot tying procedure on tissue phantom was recorded. (b) Manual annotation was perfromed by placing key points along the thread centerline. (c) The gradient map is displayed with false color. The green region is closed to the tail end of thread while the blue region is closed to the needle end. (d) The overlap map with green, blue, red color to indicate non-overlaped area, overlaped area and background}\label{fig::data}
\end{figure}

The remainder of this paper is organized as follows: In Section~\ref{sec::data}, we briefly present the procedure for data collection and ground truth annotation in our work. The problem formulation and the main workflow of the proposed method are presented in Section~\ref{sec::method}. Empirical experiments and studies on real datasets are conducted in Section~\ref{sec::exp}. Finally, the conclusions and future works are presented in Section~\ref{sec::conclusion}.

\section{Data Collection}\label{sec::data}

As shown in Fig.\ref{fig::data}(a), we firstly introduce the data collection and ground truth labeling in our work. The images used for training our neural network was collected in a knot tying procedure. This procedure was conducted by using the needle holder to wind the suture thread around the forceps and then used the forceps to pull the tail of the suture through the loop. The user can either make left over right knot or right over left knot depending on the direction to wind the suture thread on the forceps. We recorded the full procedure as multiple image frames and divide them into training and testing data. Given the original image frame which contains suture threads, the user was asked to annotate key points along the thread from the head to the tail as shown in Fig.\ref{fig::data}(b). During the labeling process, the user also indicated whether the key points are occluded. Based on the set of key points, a B-spline curve is estimated that covers the curvilinear structure of the suture thread. We then uniformly sampled the points from the curve to generate the gradient map shown in Fig.\ref{fig::data}(c). In the gradient map, the value of points on the thread is matched with the spline parameter which is within the range the range of $[0,1]$. The segments labeled as occluded was rendered first and the other segments were rendered on top the occluded segments. The brighter region is closed to the tail end of thread while the lighter region is closed to the needle end. Besides the \textit{gradient map}, we also generate an \textit{overlap map}, which is illustrated in Fig.\ref{fig::data}(d), to indicate the self-occlusions of the thread. There are three classes of points in overlap map including the non-overlapped points in green, overlapped points in blue and background points in red.

\section{METHODOLOGY}\label{sec::method}

Fig.\ref{fig::workflow} illustrates the main framework of the proposed method. The system takes a color image as input and produces the curve reconstruction of the thread as output. Firstly, the input image $X$ is transformed into the intermediate feature $F$ by an encoder network. Then the intermediate feature is used to predict the initial gradient map $G$ and the overlap map $O$. The fusion network takes the original image, initial gradient map and overlap map as input to generate the refined gradient map $\bar{G}$ which is finally parsed by curvilinear structure detector to generate the final curvilinear estimation of surgical suture . 

\subsection{Multi-Stage Deep Networks}

The image $X$ is first analyzed by a convolutional network $f_1(\cdot)$, generating a set of feature maps $F=f_1(X) $ as input to the first stage of each branch. At the first stage, the network produces the initial estimation of gradient map $G=\phi_1(F)$ and the overlapping map $O=\eta(F)$ where $\phi_1(\cdot)$ and $\eta(\cdot)$ are multi-layer decoders consisting of deconvolutional filters. For the gradient map estimation, we use an $L_1$ loss to minimize the difference between the estimated map and the ground truth. 

\begin{equation}
L_G(G)=\sum_k\sum_{i=1}^{w}\sum_{j=1}^{h}\|G_k(i,j)-G^*_k(i,j)\|
\end{equation}


where $G^*$ is the ground truth of gradient map. Although the $L_2$ loss performs well on most of general regression tasks, it does not handle very well the self-occlusion area, this perhaps is induced by different 3D shape may have similar 2D observations.  These regions do not govern large portions of the whole images compared with the non-overlapping ones. Since $L_21$ loss accumulates of pixel-wise errors, it cannot focus more on the overlapping regions. Therefore, we introduce the overlap map $O$ as a guidance to refine the initial estimation. The prediction of overlap map is modeled as a classification problem which distinguishes the backgrounds, the overlapping regions and the non-overlapping regions. We use the weighted cross-entropy loss as follows:

\begin{equation}
L_O(O)=-\sum_k\sum_{i=1}^{w}\sum_{j=1}^{h}w_k(i,j)O^*_k(i,j)logO_k(i,j)
\end{equation}
where $O^*$ is the ground truth of overlap map and $w_k$ is the sample weight. In order to focus more on the overlapping region, the corresponding weight are set to be larger than the other regions.

After the initial estimation, the gradient map and overlap map are fused with the original image in the form of multichannel data $Y=[X;O;G]$ as input to the second stage prediction. We still follow the encoder-decoder architecture as shown in Fig.\ref{fig::workflow} to generate the refined gradient map $\bar{G}=\phi_2(f_2(Y))$ and minimize the errors $L_{\bar{G}}$ between refined gradient map and ground truth based on $L_2$ loss. Therefore, the overall loss of the multistage deep networks is:

\begin{equation}
L=L_O+L_G+L_{\bar{G}}
\end{equation}

\begin{figure}[!b]
\centering
\includegraphics[width=0.48\textwidth]{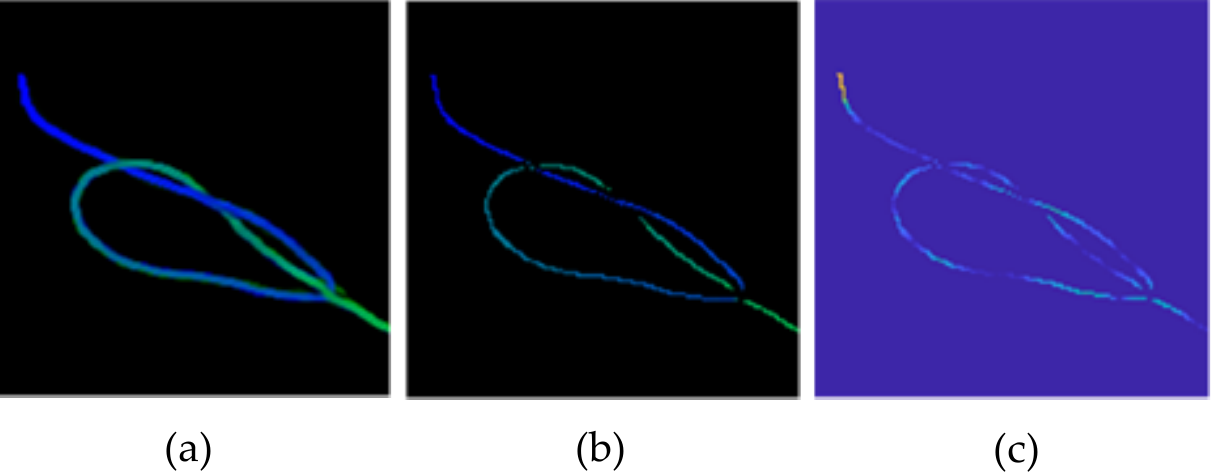}
\caption{(a) The original gradient map (b) The thin gradient map with line width reduced to 1 pixel (c) The polarity map in which pixels near an endpoint has larger value and tend to be yellow}\label{fig::thread_thin}
\end{figure}

\subsection{Thread Reconstruction}
After the gradient map is obtained, the proposed method conducts curve reconstruction by fitting a spline to the centerline of the thread. Without knowing the order of points on the centerline, it is a NP-Hard problem to fit a 2D spline when the thread is under self-intersection or occlusion. The thread gradient map not only segments the thread out from a raw image, but also provides thread structural information. One straightforward way to reconstruct the thread is to search and collect points from one endpoint of thread to the other end by following, e.g. the gradient descending direction. However, this method cannot work robustly, because if looked at locally the gradient map as shown in Fig.~\ref{fig::thread_thin}(a), the line structure tends to have a blur edge, making the gradient value inaccurate on the border area. In addition, the intersection area tends to be contaminated by noise. If the searching method select a wrong turn at one crossing point, either it cannot reach the destination (the other endpoint), or some parts of the gradient map may not be travelled. To have a robust thread reconstruction, we take a multi-stage method which searches the curve segments locally and later links them according to their average intensity value. The method contains following four key steps: 

\begin{figure}[!t]
\centering
\includegraphics[width=0.48\textwidth]{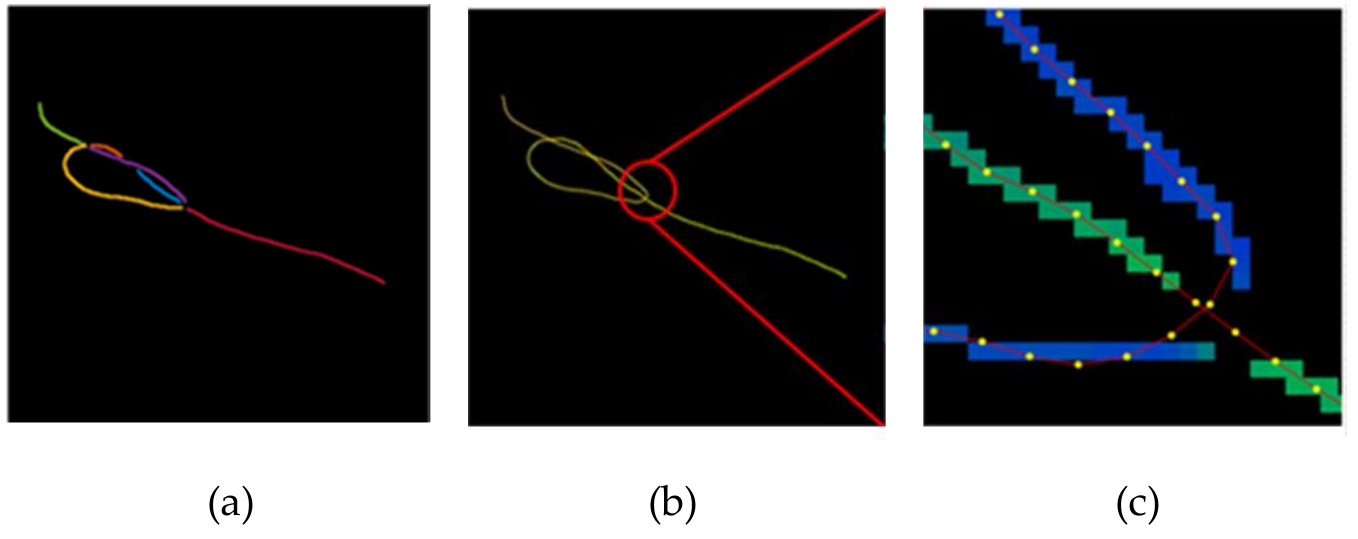}
\caption{(a) The detected curve segments represented by different colors (b) The spline fitting result (c) Zoom in the selected area, the yellow points are evenly sampled using the spline parameters}\label{fig::search}
\end{figure}

(1)	Pixels on the centerline of the gradient map is extracted as Fig.~\ref{fig::thread_thin}(b) by using the method proposed by Steger \cite{steger1998unbiased}. This step makes the later curve search easier by removing the noise on thread border and intersection area. In this step in the orientation (indicates the line direction) of each pixel is also calculated as a by-product. There are three parameters used by the centerline points extraction method. $\sigma$: It depends on the line width ($\approx$4 pixel in our case). Lower Threshold $t_l$: Line points with a response smaller as this threshold are rejected. Upper Threshold $t_u$: Line points with a response larger as this threshold are accepted.

\begin{equation}
\sigma=\frac{w}{2\sqrt{3}}+0.5\label{eq::w}
\end{equation}
where $w$ is the width of thread.

(2) The endpoints of the curvilinear structure are found using a method inspired by~\cite{tong2004first}. A polarity map Fig.~\ref{fig::thread_thin}(c) is built for the thin gradient map.  The pixel on endpoint tends to have all its neighbour pixels located at same direction and then its polarity value is higher than the other pixels. The polarity value $p$ for a pixel $\mathbf{x}$ is calculated as:

\begin{eqnarray}
\begin{split}
&p=\|\frac{1}{M}(\sum_{i=1}^M(\mathbf{x_i}-\mathbf{x})\mathbf{n})\|\\
&\|\mathbf{x_i}-\mathbf{x}\|<t_d,\|v_i-v\|<t_v\label{eq::search}
\end{split}
\end{eqnarray}

where $\mathbf{n}$ is the orientation of the pixel, $\mathbf{x_i},i=1,\ldots,M$ are neighbour pixels of $\mathbf{x}$ limited by distance threshold $t_d$ and intensity threshold $t_d$.

Start from the most salient endpoint (highest value in polarity map), we use region growing method to collect points along the line. Only those neighbouring points within the threshold $t_d $ , $t_v$ are collected.  Once a point is collected, it should be removed from the gradient map. The growing is stopped when no nearby points within the threshold $t_v$ ,$t_d$, and then we calculated the polarity map again and find a new salient endpoint. The process repeats until no points remain in the gradient map (The searching process is shown in video supplied). 

(3)	Step 2 will find multiple curve segments from the gradient map as shown in Fig.~\ref{fig::search}(a). The points in each curve may be saved either in intensity ascending or descending order. Because the intensity value is contained by noise, to determine whether it is ascending or descending, we fit a line to the intensity values, if the line has a negative slop, that means it is saved in descending order and vice versa. At last, the order of all the curve segments are found by calculating their average intensity value, and then we combined all the curve segments following this order.

(4)	The last step is to fit a cubic spline to each dimension of the ordered points separately. This step results in a curve with unwanted oscillations as shown in Fig.~\ref{fig::search}(b) and Fig.~\ref{fig::search}(c).

\section{IMPLEMENTATION}
\textbf{Deep Neural Networks} We adapt the architecture for our encoder networks from He et al. \cite{he2016deep} who have shown impressive results for image classification tasks. The fully connected layers are removed which makes the networks can cope with arbitrary sizes of input images. The decoders for gradient map and overlap map are based on the UNet~\cite{ronneberger2015u} which introduces the skip connection between down-sampling and up-sampling layers. We use the Adam solver~\cite{kingma2014adam} with a batch size of 1. The Pytorch\footnote{https://github.com/pytorch/pytorch} framework is adopted to implement the deep convolution neural networks and the experiment platform is a workstation with Xeon E5-2630 and NVIDIA GeForce Titan Xp. 

Unlike the tasks of segmentation, the output of the proposed method requires dense and continuous estimation of gradient for thread reconstruction. Due to the limitations of neural networks, the boundary of thread cannot be always sharp which is normally observed with the gradual changes between foreground and background. In order to improve the  robustness of estimation, we adopt two neural networks with the identical architecture. Besides the networks which generates the normal gradient map, the conjugated gradient map, whose starting point and end points are inverted, is obtained by another network simultaneously. We also add the constraint that the sum of outputs from two networks should equal to the binary mask of thread in the image. 

\begin{figure}[!t]
\centering
\includegraphics[width=0.48\textwidth]{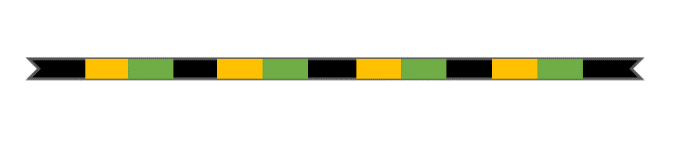}
\caption{Marked thread with repeated black, yellow and red colors.}\label{fig::data_marker}
\end{figure}

\textbf{Thread Reconstruction} The advantage of using conjugated gradient maps is that we can use it as prior information to remove the noise on the gradient map. For a pixel, If the sum of the intensity values in both gradient map is not closed to the binary mask, then it can be considered as noise and then removed from the gradient map. The centerline points extraction method works only on grayscale image. we simply sum the corresponding values in each gradient map and covert it to grayscale image. Furthermore, segments found may only contain several pixels. When linking the curve segments into a single curve, these very short segments are not considered by setting a threshold $t_c$ which is the minimal number of pixels in each curve. In this paper, we set the thread width $w=4$ in Eq.(\ref{eq::w}). The thresholds for centerline extraction in Eq.(\ref{eq::search}) are set where $t_l=0.039$ and $t_u=0.196$. For parameters in curve search, we set $t_d=2$, $t_v=0.1$ and $t_c=14$. These parameters are tuned by empirical evaluations.

\section{EXPERIMENTS}
\subsection{Experimental Settings}\label{sec::exp}
In this section, we conduct the experimental validation
to evaluate the performance of the suture detection
method. As shown in Fig.\ref{fig::data_marker}, two types of threads are used
in the experiments including the threads in single color and the threads colored with directional pattern as Fig.6. Some suture uses this kind of pattern to enhance visibility and provides distinguishing features for suture management. We assume this kind of directional marker can also remove the ambiguity (cannot differentiate the head and tail) in learning the gradient map. A test was performed to see whether it increase our suture detection compared with single color thread. 
 
The image we use is captured in size of $ 1280 \times 720$ and resized to
$512 \times 384$ for efficient calculation. For both dataset, we
collect three videos for training and one video for testing.
As a result, the number of training data is 2000 and the
testing data is 1000. The forward network, which generates
the initial gradient map, is set as a baseline to validate the
performance of the proposed method.

\subsection{Quantitative Evaluation}
\begin{table}[!t]
\centering
\caption{Prediction error of the proposed method.}\label{tab::quat}
\begin{tabular}{ccccc}
\hline
\hline
&\multicolumn{2}{c}{Thread with markers} &\multicolumn{2}{c}{Thread without markers} \\
\hline
 &$\text{PSNR}$&$\text{OTTP}$&$\text{PSNR}$&$\text{OTTP}$\\
 DMSD-S & 31.82$\pm$0.73& 1.37$\pm$2.25& 31.48$\pm$0.99& 9.69$\pm$27.65\\
 DMSD-O & 31.91$\pm$0.70& 1.33$\pm$1.09& 31.61$\pm$0.95& 4.27$\pm$7.48\\
\hline
\hline
\end{tabular}
\end{table}

\begin{table*}[!t]
\centering
\caption{Prediction errors of different thread part and occlusion status with OTTP.}\label{tab::quat_part}
\begin{tabular}{ccccccc}
 \hline
 \hline
 &\multicolumn{3}{c}{Thread with markers} &\multicolumn{3}{c}{Thread w/o markers} \\
 \hline
 &No occlusion&Self-intersection&Tools-occlusion&No occlusion&Self-intersection&Tools-occlusion\\
 Thread Body & $1.06\pm0.23$&$	1.14	\pm0.16$&	$1.09\pm	0.26$&$1.17\pm	0.20$&	$1.12	\pm0.29$&$	1.31	\pm0.62$\\
 Thread Needle End &$5.32\pm	3.40$	&$8.71	\pm6.70	$&$6.19\pm	4.48$&$4.91	\pm4.21$&$	7.94\pm	4.83	$&$12.91\pm	16.62$\\
 Thread Tail End &$2.87	\pm1.98$&$	3.50\pm	3.51	$&$3.94\pm	3.30 $&$3.42	\pm3.61	$&$8.10\pm	6.22	$&$12.15	\pm25.68$\\
 \hline
 \hline
\end{tabular}
\end{table*}
In order to present the quantitative evaluation of the proposed method, we report the accuracy based on the gradient map and spline of thread by two criterion including peak-signal-to-noise ratio (PSNR) and overall thread tracking precision~\cite{wang2009robust}.
\begin{itemize}
\item \textbf{Peak-Signal-to-Noise Ratio (PSNR)} has been widely adopted as evaluation metric in image filtering and reconstruction. In the experiment, we calculate \text{PSNR} between the ground truth $G^*$ and estimated gradient map $\bar{G}$. The mean value of PSNR over the testing data is presented. The larger value of PSNR indicates the better accuracy.

\item \textbf{Overall Thread Tracking Precision (\text{OTTP})} is defined as the average of shortest distances from points on a reconstructed thread $S(\theta)$ to the corresponding ground truth $S^{*}$. Such a precision describes how close a result to the ground truth as follows:
\begin{eqnarray}
\begin{split}
&\text{OTTP}=\frac{1}{m}{\sum_{i=1}^m\frac{1}{l_i}\sum_{j=1}^{l_i}d(S_j(\theta),S^*)}\\
&d(S_j(\theta),S^*)=\min_{k} \|S_j(\theta)-S^*_k\|
\end{split}
\end{eqnarray}
\end{itemize}
where $m$ is the number of testing data and $l_i$ is the number of points sampled from the spline. The smaller value of OTTP indicates the better accuracy.

\begin{figure*}[!t]
\centering
\includegraphics[width=0.9\textwidth]{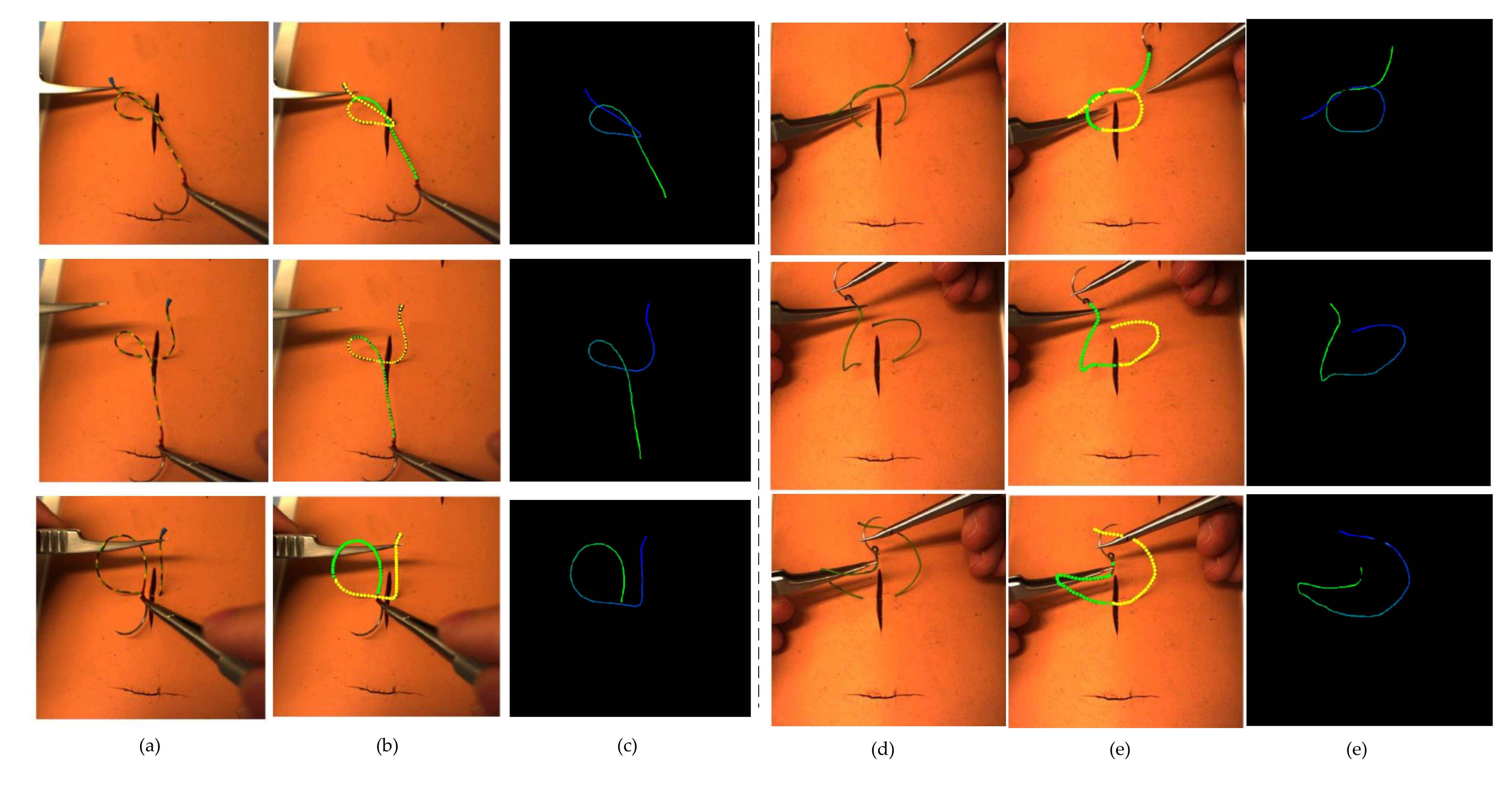}
\caption{Qualitative evaluation of the proposed method.(a,d) The orginal images of thread with/withour markers (b,e) The result of curve reconstruction: to visualize overlapping, use two colors. (c,d) The gradient maps for each case.}\label{fig::vis_all}
\end{figure*}

\begin{figure*}[!t]
\centering
\includegraphics[width=0.9\textwidth]{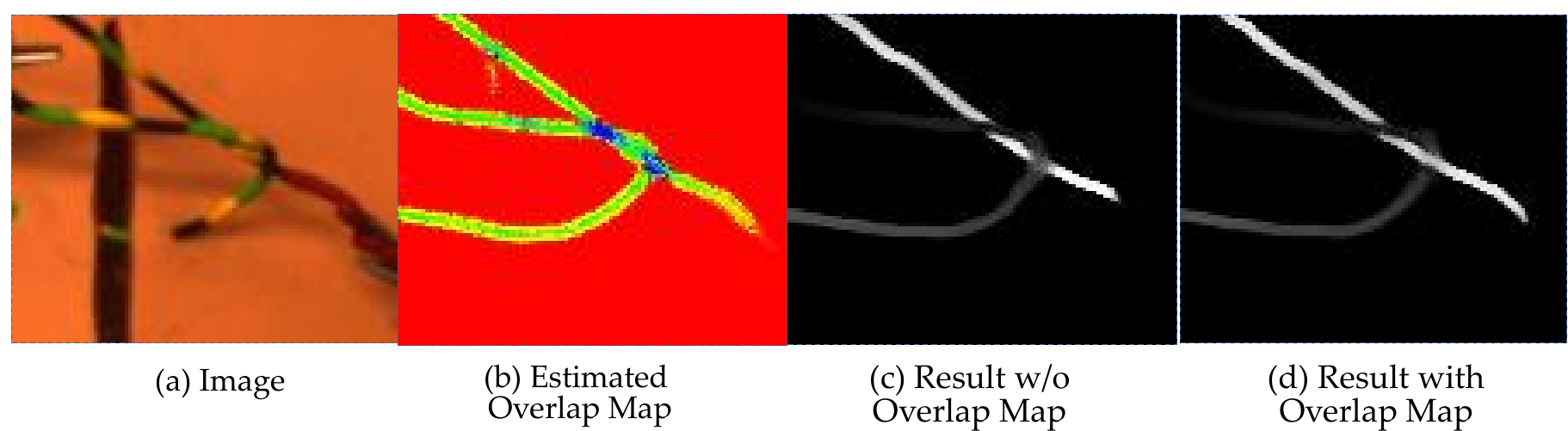}
\caption{Qualitative result of self-occlusion.}\label{fig::vis_ablation}
\end{figure*}

The quantitative result of the proposed method is reported in Table~\ref{tab::quat} where DMSD-S is the baseline and DMSD-O is the proposed method. Compared to the baseline, the proposed method can achieve higher accuracy on both PSNR and OTTP metrics. Although the improvement on thread with markers is relatively slight, the proposed method obtains smaller standard deviation which indicates the good stability and robustness. For the thread without markers, the improvement over the baseline is significant. Since PSNR only measures the image-level errors between the estimated data and ground truth, we achieve relatively slight improvement over the baseline. However, as mentioned in Section~\ref{sec::method}, small errors on the overlapped region of gradient map will lead to totally different splines. Therefore, the proposed method achieves much smaller errors on OTTP metric compared to the baseline. 

To better evaluate our method on thread reconstruction, the recorded frames were divided into three groups: (1) thread is without occlusion and entanglement, (2) the thread intersects itself for a finite length, e.g. overhand knot. (3) part of thread is occluded by the tool. The typical case for each group is shown as Fig.\ref{fig::vis_all}(a) and Fig.\ref{fig::vis_all}(d). Besides the overall thread centerline reconstruction accuracy, we also report the accuracy on the two endpoints of thread. As shown in Table~\ref{tab::quat_part}, finite self-intersection during overhand knotting does not have severe influence on the accuracy if the whole thread is clearly visible. On the contrary, the occlusion, because of thread and instruments interaction, will reduce the accuracy and increase missing detection rate (the standard deviation increases because fail in detection), especially in no marker case. It is also found that the two endpoints has lower accuracy than the overall accuracy. This is because the highest intensity value in the intensity map is always located within a neighbour of the endpoint. Comparing the two types of thread, the centerline reconstruction accuracy of the marker thread is consistently higher than thread with no marker. It verifies our assumption that the directional pattern design provides better visibility and distinguishing features for thread recognition. 

Finally, we also test the running time of the proposed method. The proposed method can generate the reconstruction of thread at 15fps on our platform. 

\subsection{Qualitative Evaluation}

\begin{figure*}[!t]
\centering
\includegraphics[width=0.9\textwidth]{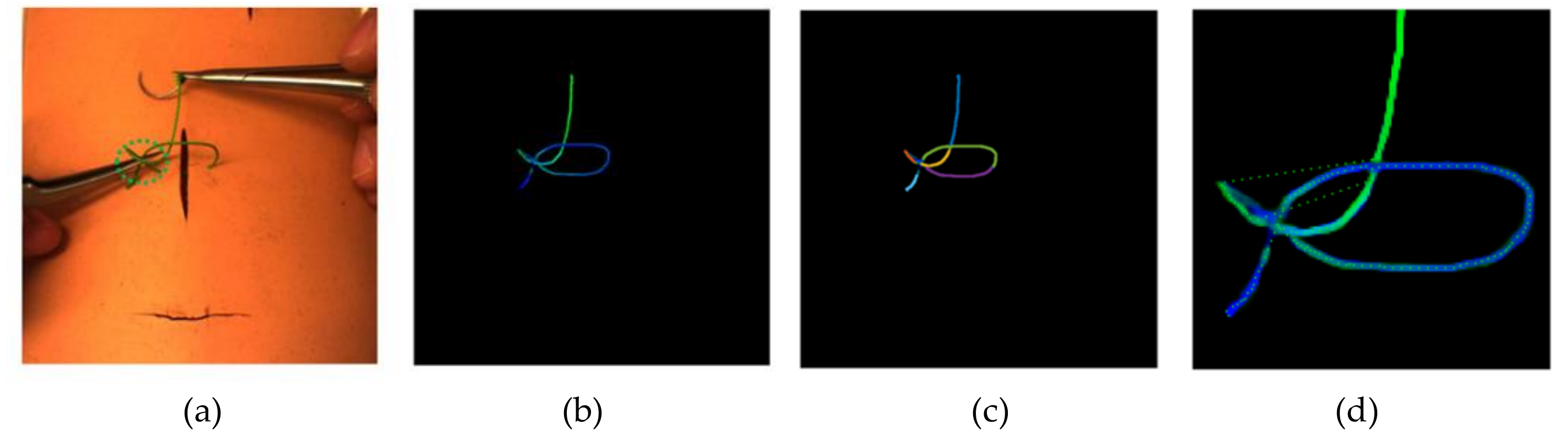}
\caption{A fail detection the thread with no markers. (a) Original image (b)Gradient map
(c) Found curve segments (d) The reconstructed curve is drawn on the gradient map.
}\label{fig::failure_no_marker}
\end{figure*}

 \begin{figure*}[!t]
\centering
\includegraphics[width=0.9\textwidth]{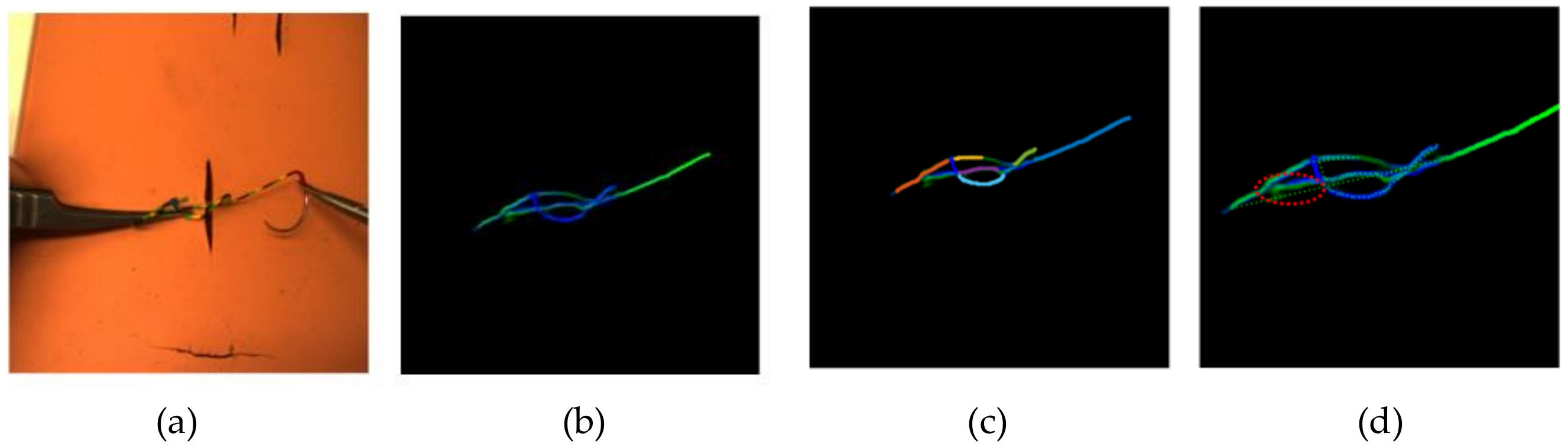}
\caption{A fail detection the thread with markers. (a) Original image (b) Gradient map
(c) Found curve segments (d) The reconstructed curve is drawn on the gradient map. 
}\label{fig::failure_marker}
\end{figure*}
 
In addition to the quantitative evaluations, the global result of six cases on two datasets are presented in Fig.~\ref{fig::vis_all}. For each dataset, we provide the raw image, the reconstructed spline and the gradient map. The reconstructed spline is projected onto the raw image. In order to present visualization, points which are closed to the needle end is colored in yellow while the points closed to tail end is colored in green. According to Fig.~\ref{fig::vis_all}(b) and Fig.~\ref{fig::vis_all}(e), the curvilinear structure of thread  is properly reconstructed even with complicated self-occlusion. From the perspective of gradient map, the trend of intensity matches the real structure of thread. Although the gradient map is not complete due to the tools-occlusion in some cases, the local curve search algorithm can still fill the gap. Besides the global result, we also present the local improvement contributed by overlap map in Fig.~\ref{fig::vis_ablation}. As mentioned in Section ~\ref{sec::method}, the error of the self-occlusion region is relatively small compared to the global image, which cannot attract higher attention from baseline method (DMSD-S). However, the proposed method introduces the overlap map to predict the self-intersection region as guidance to refine the initial gradient map. The comparison between Fig.~\ref{fig::vis_ablation}(c) and Fig.~\ref{fig::vis_ablation}(d) demonstrates the improvement of the proposed method. 

Two sets of typical failure cases are presented for further analysis. In complicated thread manipulation as shown in Fig.\ref{fig::failure_no_marker}, our method fails but human can reconstruct the thread curve.  In the green circle in Fig.\ref{fig::failure_no_marker}(a), three layers of thread segments overlap, in our ground truth we did not label this situation (the self-occlusion only involves two segments), so the neural network is incapable of producing a good gradient map in this situation (Fig.\ref{fig::failure_no_marker}(b)). Even though we can extract multiple curves (Fig.\ref{fig::failure_no_marker}(c)) from the images, the order of these curve according to their mean intensity value is not correct, so the spline fitting is not correct (Fig.\ref{fig::failure_no_marker}(d)).

Those cases fail in marked thread detection as shown in Fig.\ref{fig::failure_marker} are mainly because its curve reconstruction is too difficult even for human. e.g Fig.\ref{fig::failure_marker}(a). In this case, the thread is severely entangled and occluded, making the gradient map very inaccuracy in the entanglement area (Fig.\ref{fig::failure_marker} (b)).  In a real application requires thread detection, usually it does not need vision to work on this situation. For example, just pulling the two ends of thread without looking can still finish the knot. And for knot type recognition task, the knot type can be recognized by understanding previous simple scene.

\section{DISCUSSION AND CONCLUSION}\label{sec::conclusion}

In this paper, we propose a multi-stage framework for suture detection based on deep learning. Instead of the individual steps segmentation and thread-modelling in previous work, the proposed method tends to directly predict the curvilinear structural information of the thread via deep convolutional neural networks. We firstly estimate the dense prediction of curvilinear structure and the overlapping status of the thread by a two-branched CNN. The initial estimation is further fused with the original visual observation and fed into the second stage CNN to generate the refined prediction. Based on the gradient map, multiple segments of the thread are extracted and linked to form the whole thread with organized points located on its centerline using a curvilinear structure detector. Our method does not rely on any model for optimization, a simple spline is used for smoothing the curve reconstruction at the last step.  Experiment on two types of surgical thread indicates that the proposed detection method can achieve good performance when no occlusion or finite self-occlusion (e.g. overhand knot). For the tools-occlusion cases, the detection deteriorates for both type of threads, however, human level detection performance can be achieved for the thread with directional markers. The main limitation of the proposed method is that the sequential information is not adopted for thread reconstruction. In the future work, we plan to integrate the time-specific constraints into our framework by recurrent neural networks (RNN).

\bibliographystyle{IEEEtran} 
\bibliography{IEEEabrv,./IEEEexample}

\begin{thebibliography}{10}
\providecommand{\url}[1]{#1}
\csname url@samestyle\endcsname
\providecommand{\newblock}{\relax}
\providecommand{\bibinfo}[2]{#2}
\providecommand{\BIBentrySTDinterwordspacing}{\spaceskip=0pt\relax}
\providecommand{\BIBentryALTinterwordstretchfactor}{4}
\providecommand{\BIBentryALTinterwordspacing}{\spaceskip=\fontdimen2\font plus
\BIBentryALTinterwordstretchfactor\fontdimen3\font minus
  \fontdimen4\font\relax}
\providecommand{\BIBforeignlanguage}[2]{{%
\expandafter\ifx\csname l@#1\endcsname\relax
\typeout{** WARNING: IEEEtran.bst: No hyphenation pattern has been}%
\typeout{** loaded for the language `#1'. Using the pattern for}%
\typeout{** the default language instead.}%
\else
\language=\csname l@#1\endcsname
\fi
#2}}
\providecommand{\BIBdecl}{\relax}
\BIBdecl

\bibitem{jin2005automated}
X.~Jin and C.~H. Davis, ``Automated building extraction from high-resolution
  satellite imagery in urban areas using structural, contextual, and spectral
  information,'' \emph{EURASIP Journal on Advances in Signal Processing}, vol.
  2005, no.~14, p. 745309, 2005.

\bibitem{wang2009robust}
P.~Wang, T.~Chen, Y.~Zhu, W.~Zhang, S.~K. Zhou, and D.~Comaniciu, ``Robust
  guidewire tracking in fluoroscopy,'' in \emph{Computer Vision and Pattern
  Recognition, 2009. CVPR 2009. IEEE Conference on}.\hskip 1em plus 0.5em minus
  0.4em\relax IEEE, 2009, pp. 691--698.

\bibitem{hoover2000locating}
A.~Hoover, V.~Kouznetsova, and M.~Goldbaum, ``Locating blood vessels in retinal
  images by piecewise threshold probing of a matched filter response,''
  \emph{IEEE Transactions on Medical imaging}, vol.~19, no.~3, pp. 203--210,
  2000.

\bibitem{schulman2013tracking}
J.~Schulman, A.~Lee, J.~Ho, and P.~Abbeel, ``Tracking deformable objects with
  point clouds,'' in \emph{Robotics and Automation (ICRA), 2013 IEEE
  International Conference on}.\hskip 1em plus 0.5em minus 0.4em\relax IEEE,
  2013, pp. 1130--1137.

\bibitem{schulman2016learning}
J.~Schulman, J.~Ho, C.~Lee, and P.~Abbeel, ``Learning from demonstrations
  through the use of non-rigid registration,'' in \emph{Robotics
  Research}.\hskip 1em plus 0.5em minus 0.4em\relax Springer, 2016, pp.
  339--354.

\bibitem{lui2013tangled}
W.~H. Lui and A.~Saxena, ``Tangled: Learning to untangle ropes with rgb-d
  perception,'' in \emph{Intelligent Robots and Systems (IROS), 2013 IEEE/RSJ
  International Conference on}.\hskip 1em plus 0.5em minus 0.4em\relax IEEE,
  2013, pp. 837--844.

\bibitem{javdani2011modeling}
S.~Javdani, S.~Tandon, J.~Tang, J.~F. O'Brien, and P.~Abbeel, ``Modeling and
  perception of deformable one-dimensional objects,'' in \emph{Robotics and
  Automation (ICRA), 2011 IEEE International Conference on}.\hskip 1em plus
  0.5em minus 0.4em\relax IEEE, 2011, pp. 1607--1614.

\bibitem{ambrosini2017fully}
P.~Ambrosini, D.~Ruijters, W.~J. Niessen, A.~Moelker, and T.~van Walsum,
  ``Fully automatic and real-time catheter segmentation in x-ray fluoroscopy,''
  in \emph{International Conference on Medical Image Computing and
  Computer-Assisted Intervention}.\hskip 1em plus 0.5em minus 0.4em\relax
  Springer, 2017, pp. 577--585.

\bibitem{padoy2012deformable}
N.~Padoy and G.~D. Hager, ``Deformable tracking of textured curvilinear
  objects.'' in \emph{BMVC}, 2012, pp. 1--11.

\bibitem{jackson2015automatic}
R.~C. Jackson, R.~Yuan, D.-L. Chow, W.~Newman, and M.~C.
  {\c{C}}avu{\c{s}}o{\u{g}}lu, ``Automatic initialization and dynamic tracking
  of surgical suture threads,'' in \emph{Robotics and Automation (ICRA), 2015
  IEEE International Conference on}.\hskip 1em plus 0.5em minus 0.4em\relax
  IEEE, 2015, pp. 4710--4716.

\bibitem{jackson2017real}
R.~C. Jackson, R.~Yuan, D.-L. Chow, W.~S. Newman, and M.~C.
  {\c{C}}avu{\c{s}}o{\u{g}}glu, ``Real-time visual tracking of dynamic surgical
  suture threads,'' \emph{IEEE Transactions on Automation Science and
  Engineering}, 2017.

\bibitem{steger1998unbiased}
C.~Steger, ``An unbiased detector of curvilinear structures,'' \emph{IEEE
  Transactions on pattern analysis and machine intelligence}, vol.~20, no.~2,
  pp. 113--125, 1998.

\bibitem{strokina2013detection}
N.~Strokina, T.~Kurakina, T.~Eerola, L.~Lensu, and H.~K{\"a}lvi{\"a}inen,
  ``Detection of curvilinear structures by tensor voting applied to fiber
  characterization,'' in \emph{Scandinavian Conference on Image
  Analysis}.\hskip 1em plus 0.5em minus 0.4em\relax Springer, 2013, pp. 22--33.

\bibitem{fu2016retinal}
H.~Fu, Y.~Xu, D.~W.~K. Wong, and J.~Liu, ``Retinal vessel segmentation via deep
  learning network and fully-connected conditional random fields,'' in
  \emph{Biomedical Imaging (ISBI), 2016 IEEE 13th International Symposium
  on}.\hskip 1em plus 0.5em minus 0.4em\relax IEEE, 2016, pp. 698--701.

\bibitem{charbonnier2017improving}
J.-P. Charbonnier, E.~M. Van~Rikxoort, A.~A. Setio, C.~M. Schaefer-Prokop,
  B.~van Ginneken, and F.~Ciompi, ``Improving airway segmentation in computed
  tomography using leak detection with convolutional networks,'' \emph{Medical
  image analysis}, vol.~36, pp. 52--60, 2017.

\bibitem{tong2004first}
W.-S. Tong, C.-K. Tang, P.~Mordohai, and G.~Medioni, ``First order augmentation
  to tensor voting for boundary inference and multiscale analysis in 3d,''
  \emph{IEEE transactions on pattern analysis and machine intelligence},
  vol.~26, no.~5, pp. 594--611, 2004.

\bibitem{he2016deep}
K.~He, X.~Zhang, S.~Ren, and J.~Sun, ``Deep residual learning for image
  recognition,'' in \emph{Proceedings of the IEEE conference on computer vision
  and pattern recognition}, 2016, pp. 770--778.

\bibitem{ronneberger2015u}
O.~Ronneberger, P.~Fischer, and T.~Brox, ``U-net: Convolutional networks for
  biomedical image segmentation,'' in \emph{International Conference on Medical
  Image Computing and Computer-Assisted Intervention}.\hskip 1em plus 0.5em
  minus 0.4em\relax Springer, 2015, pp. 234--241.

\bibitem{kingma2014adam}
D.~Kingma and J.~Ba, ``Adam: A method for stochastic optimization,''
  \emph{arXiv preprint arXiv:1412.6980}, 2014.

\end{thebibliography}


\end{document}